\documentclass[sigconf, nonacm]{acmart}

\usepackage{multirow}
\usepackage{adjustbox}
\usepackage{algorithm}
\usepackage{algpseudocode}

\AtBeginDocument{%
  }

\begin{document}

\title{SURGENT: A Surgical Multi-Agent Assistance System Across the Perioperative Workflow}

\author{Dongsheng Shi}
\affiliation{%
  \institution{East China Normal University}
  \city{Shanghai}
  \country{China}
}
\email{dongsheng@stu.ecnu.edu.cn}

\author{Yue Li}
\affiliation{%
  \institution{East China Normal University}
  \city{Shanghai}
  \country{China}
}
\email{yue\_li@stu.ecnu.edu.cn}

\author{Xin Yi}
\affiliation{%
  \institution{East China Normal University}
  \city{Shanghai}
  \country{China}
}
\email{xinyi@stu.ecnu.edu.cn}

\author{Yongyi Cui}
\affiliation{%
  \institution{East China Normal University}
  \city{Shanghai}
  \country{China}
}
\email{yycui@stu.ecnu.edu.cn}

\author{Huawei Feng}
\affiliation{%
  \institution{East China Normal University} 
  \city{Shanghai} 
  \country{China} 
} 
\email{hwfeng@stu.ecnu.edu.cn}

\author{Linlin Wang} 
\authornote{Corresponding author.} 
\affiliation{%
  \institution{East China Normal University} 
  \city{Shanghai} 
  \country{China} 
} 
\affiliation{%
  \institution{City University of Hong Kong} 
  \city{Hong Kong SAR} 
  \country{China} 
}
\email{llwang@cs.ecnu.edu.cn}

\begin{abstract}

The intricate nature of modern surgical care necessitates intelligent systems that can synthesize extensive patient records, support collaborative decision-making, and provide transparent, auditable reasoning across the entire perioperative workflow. Although web-based Large Language Models (LLMs) possess advanced reasoning capabilities, they are ill-equipped for surgical applications due to critical limitations: input length constraints, incomplete memory management, and limited traceability. 
To address this issue, 
we present \textbf{SURGENT}, a \underline{\textbf{surg}}ical multi-ag\underline{\textbf{ent}} assistance system that combines a Tree-of-Thought planner, multi-department collaboration agents, and retrieval-augmented reasoning with clinical guidelines and biomedical literature. SURGENT features a novel memory design that manages both long-term patient histories and short-term working summaries, enabling more complete, contextualized, and consistent reasoning. Experimental evaluations across five key perioperative tasks—case analysis, surgical plan simulation, safety monitoring, complication risk assessment, and rehabilitation guidance—show that SURGENT outperforms baseline LLMs and existing medical multi-agent frameworks, yielding recommendations more closely aligned with patient histories. Ablation studies further highlight the advantage of DeepSeek as a locally deployable backbone model, enabling privacy-preserving deployment without reliance on centralized services. These results position SURGENT as a practical and trustworthy advancement toward intelligent, equitable, and secure surgical assistance systems. 
\end{abstract}





\maketitle

\section{Introduction}

In recent years, the demand for surgical services has been rapidly increasing due to population growth, aging, and the rising prevalence of complex diseases \cite{gruber2025china, li2020effect}. Recent work has pointed out that the application of large language models (LLMs) in the surgical domain is still in an exploratory and developmental stage, facing challenges related to data security, ethical concerns, and technical limitations \cite{Zheng2024AIinMedicine}. Although several hundred thousand registered surgeons serve a vast population, their distribution is highly uneven, with most concentrated in large urban hospitals, leaving many county and rural hospitals under-resourced \cite{yu2021physician, wang2020disequilibrium}. These pressures highlight the urgent need for intelligent surgical assistance systems that can support decision-making, improve consistency, and leverage extensive patient records, particularly in under-resourced regions \cite{guni2024artificial, balakrishnan2025artificial, sheikh2019artificial}.

\begin{figure}[h]
  \includegraphics[width=\columnwidth]{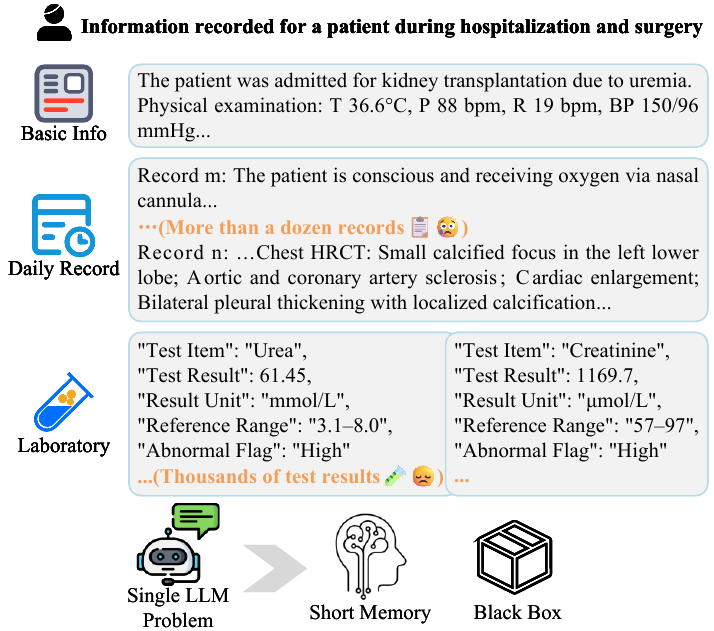}
  \caption{Challenges of Using a Single LLM in Analyzing Extensive Medical Records and Laboratory Test Results for Surgical Assistance.}
  \label{fig:intro}
\end{figure}

LLMs have demonstrated strong performance across a wide range of domains, ranging from general-purpose reasoning to domain-specific decision support
\cite{shi2026benchmarking, li2025hierarchical, yi2025unified, yi2025latent}.
As these capabilities are increasingly delivered through online Web platforms, advanced reasoning services have become accessible to users without requiring specialized infrastructure or local deployment. This trend has further lowered the barrier to accessing medical knowledge and decision-support information \cite{dennstadt2025implementing, riedemann2024path}. Patients and practitioners alike can query such systems directly through a browser, democratizing access to expert-level insights that were once largely restricted to specialized institutions \cite{platt2024public, frodl2024chatgpt}. 

Despite this improved accessibility, deploying Web-based LLMs in surgical scenarios remains substantially more challenging than using them for general medical information seeking. As shown in Figure \ref{fig:intro}, in surgical contexts, patient records are often extensive, encompassing prior diagnoses, daily medical records, lab results, and operative notes \cite{hasan2023preserving, adams2025longhealth}. Standard Web-based LLMs cannot fully ingest such long histories in a single query, which can lead to incomplete or inconsistent reasoning \cite{yan2025inftythink, hosseini2024efficient}. Moreover, the high-stakes nature of surgery demands transparent, traceable decision processes and rigorous protection of sensitive patient data—requirements that current Web-based LLM services struggle to meet \cite{mbadjeu2025large, wang2024applications, aljohani2025comprehensive}.

To address these challenges, we present SURGENT, a surgical multi-agent assistance system designed explicitly around the Web’s principles of accessibility, fairness, inclusiveness \cite{gunasekara2025systematic, microsoft_rAI}. SURGENT is equipped with a Tree-of-Thought (ToT) \cite{yao2023tree} planner and multi-department collaboration agents \cite{tang2024medagents, kim2024mdagents}, enabling transparent and coordinated decision workflows. It further integrates tool-augmented  reasoning with access to PubMed and clinical guideline retrieval, ensuring traceable and knowledge-grounded recommendations \cite{lewis2020retrieval, yang2025retrieval, gargari2025enhancing}. By effectively managing long-term medical histories and short-term working summaries, this memory design mitigates the input length constraints of conventional LLMs and ensures more complete, contextualized, consistent, and efficient reasoning throughout the perioperative workflow \cite{packer2023memgpt, zhang2025survey}.

 SURGENT is structured around the perioperative workflow, with five key tasks: preoperative \textbf{intelligent case analysis}, \textbf{personalized surgical plan simulation}, intraoperative \textbf{safety monitoring and alerting}, postoperative \textbf{complication risk assessment}, and \textbf{personalized rehabilitation guidance} \cite{zambouri2007preoperative, martinez2024perioperative}. Extensive experiments show that SURGENT outperforms baseline systems on each of these tasks, demonstrating its effectiveness in providing context-aware, reliable, and transparent surgical assistance. Deployed as a secure Web service, SURGENT enables doctors in under-resourced hospitals to access the same high-quality assistance available in leading urban centers, narrowing regional disparities in surgical care.

The main contributions are summarized as follows:
\begin{itemize}
    \item To the best of our knowledge, we present the first surgical multi-agent assistance system, which integrates a Tree-of-Thought planner, multi-department collaboration agents, and retrieval-based knowledge grounding, while explicitly maintaining both short-term and long-term patient memories.

    \item Extensive experiments demonstrate that SURGENT outperforms both single LLM baselines and existing medical multi-agent systems. The multi-agent collaborative workflow not only enhances transparency in decision-making but also leverages its memory and retrieval mechanisms to generate surgical recommendations that more accurately align with the patient’s historical medical records.

    \item Through ablation studies, we show that DeepSeek, when used as the backbone LLM, yields the best overall performance. Importantly, as an open-source and locally deployable model, DeepSeek enables privacy-preserving deployment, avoiding the data exposure risks of proprietary, centralized LLM services.
\end{itemize}

\section{Task Formulation}
In our multi-agent surgical system, we define a set of high-level task categories, each corresponding to a distinct functional objective. Let
\[
\mathcal{T} = \{ \tau_0, \tau_1, \tau_2, \tau_3, \tau_4 \}
\]
denote the set of task descriptions, with each task associated with a dedicated planner-assigned agent.

Each task is formally defined as follows:

\begin{enumerate}
    \item \textbf{Analysis} $(\tau_0)$:  
    Given patient admission records and medical history, perform intelligent case analysis, identify potential diagnostic risks, prevent missed or incorrect diagnoses, and generate comprehensive decision-support recommendations.

    \item \textbf{Surgery} $(\tau_1)$:  
    Integrate patient-specific characteristics, prior cases, and relevant clinical guidelines to generate a personalized surgical plan simulation that is rational, safe, and clinically informative.

    \item \textbf{Safety} $(\tau_2)$:  
    Monitor intraoperative risks in real time, provide safety protection and early warning mechanisms, and ensure controllability and patient safety throughout the surgical procedure.

    \item \textbf{Risk} $(\tau_3)$:  
    Based on postoperative status and clinical data, detect potential complications, predict likely adverse events, and provide preventive guidance.

    \item \textbf{Rehabilitation} $(\tau_4)$:  
    Considering postoperative recovery and individual patient differences, offer personalized rehabilitation guidance including lifestyle recommendations, rehabilitation exercises, and follow-up schedules.
\end{enumerate}

These task formulations provide the foundation for planner-guided agent assignment, memory retrieval, and aggregation, ensuring that each functional module operates under clear clinical objectives while interacting coherently with other departments. 

Specifically, the tasks span the entire surgical workflow: \texttt{analysis} and \texttt{surgery} are primarily performed \textbf{\emph{preoperatively}} to support diagnosis, case assessment, and personalized surgical planning; \texttt{safety} operates \textbf{\emph{intraoperatively}} to monitor real-time risks and ensure procedural safety; \texttt{risk} and \texttt{rehab} are \textbf{\emph{postoperative}} tasks, focusing on complication prediction, preventive interventions, and tailored rehabilitation guidance. 

\section{Methods}
\begin{figure*}[ht]
  \includegraphics[width=\textwidth]{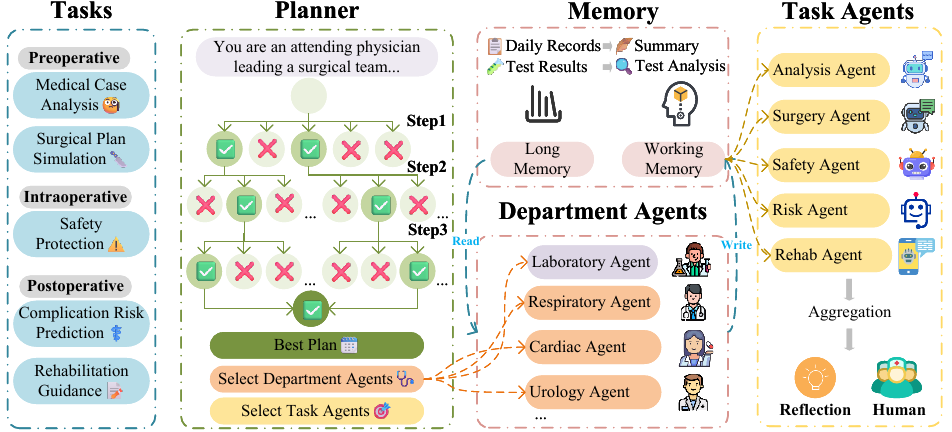}
  \caption{Diagram of our surgical multi-agent assistance system, including task definition, planner, memory, department agents, and task agents.}
  \label{fig:framework}
\end{figure*}

We present a multi-agent surgical assistance system that integrates patient-specific data, domain knowledge, and advanced reasoning mechanisms. The system comprises four core components (Figure \ref{fig:framework}): a planner agent for stepwise surgical plan generation and agent allocation (see Section \ref{subsection:planner}), a memory mechanism for managing short-term and long-term information (see Section \ref{subsection:memory}), and department agents that execute domain-specific reasoning (see Section \ref{subsection:department}). Outputs from departmental agents are consolidated through an aggregation module, which incorporates reflective reasoning and human-in-the-loop oversight to ensure clinical reliability (see Section \ref{subsection:aggregation}).

\subsection{Planner Agent}
\label{subsection:planner}
To ensure transparent and optimized surgical planning, we design a
planner agent that integrates beam search \cite{sutskever2014sequence, bahdanau2014neural} with an LLM-based evaluator.
The planner decomposes a surgical task into stepwise plans, evaluates each candidate plan under a multi-dimensional scoring rubric, and adaptively selects the most promising sequence of actions. The workflow can be divided into two major components: \emph{plan generation and search}, and \emph{multi-agent allocation}.

\noindent \textbf{Plan Generation and Evaluation.}
We formalize a candidate surgical plan as a sequence:
\begin{equation}
    P = \{p_{1}, p_{2}, \dots, p_{k}\},
\end{equation}
where each $p_{i}$ denotes one surgical step. 

To guide evaluation, we define a normalized scoring function
$S(P) \in [0,1]$ composed of five weighted criteria:
\begin{align}
S(P) = & \; \alpha_{1} \cdot f_{\text{task}}(P) 
       + \alpha_{2} \cdot f_{\text{safety}}(P) \notag \\
       & + \alpha_{3} \cdot f_{\text{logic}}(P) 
       + \alpha_{4} \cdot f_{\text{oper}}(P) \notag \\
       & + \alpha_{5} \cdot f_{\text{clarity}}(P),
\end{align}
with $\sum_{i=1}^{5} \alpha_{i} = 1$.
Specifically, the rubric is:
\begin{itemize}
    \item Task alignment $f_{\text{task}}$: whether the plan addresses the core surgical objective (0–0.30).
    \item Safety compliance $f_{\text{safety}}$: inclusion of sterile procedures, bleeding/infection control (0–0.25).
    \item Logical order $f_{\text{logic}}$: correctness of temporal dependencies between steps (0–0.20).
    \item Operability $f_{\text{oper}}$: concreteness and feasibility of steps (0–0.15).
    \item Conciseness $f_{\text{clarity}}$: clarity and brevity of instructions (0–0.10).
\end{itemize}

Given an initial root node with an empty plan, the planner explores stepwise expansions under a \emph{beam search} procedure (Algorithm \ref{alg:planner}). At each depth $d$, a set of candidates $\mathcal{C}_{d}$ is generated:
\begin{equation}
\mathcal{C}_{d} = \bigcup_{P \in \mathcal{B}_{d-1}} 
  \left\{ P \oplus p \;\middle|\; p \in M_{\text{LLM}}(P, B, \tau) \right\},
\end{equation}
where $\mathcal{B}_{d-1}$ is the beam at depth $d-1$, 
$M_{\text{LLM}}$ denotes the LLM generating $k$ novel step suggestions given the current prefix plan $P$, patient's basic information $B$ and task $\tau$, and $\oplus$ is the concatenation operator. Each candidate $P$ is then scored by $S(P)$, and the top-$b$ plans are retained:
\begin{equation}
\mathcal{B}_{d} = \text{Top-}b \left( \mathcal{C}_{d}, S(P) \right).
\end{equation}
After $D$ iterations, the highest-scoring plan $P^{*}$ is selected.

\noindent \textbf{Multi-Agent Allocation.}
Once the optimal plan $P^{*}$ is obtained, the planner assigns execution to appropriate agents at two levels:
\begin{enumerate}
    \item Task Agent Selection.  
    Given the global task description $\tau$, a meta-prompt classifier selects one agent type 
    \[
    A_{\text{task}} \in 
    \{ \texttt{analysis}, \texttt{surgery}, \texttt{safety}, \texttt{risk}, \texttt{rehab}\}.
    \]
    This ensures that the downstream reasoning is directed to the correct high-level functional module.
    \item Local Agent Selection.  
    Based on patient profile information, the planner further identifies a subset of specialized clinical agents:
    \[
    \mathcal{A}_{\text{local}} \subseteq 
    \{\texttt{cardiovascular}, \texttt{orthopedics}...\}.
    \]
    These local agents provide domain-specific expertise, enabling multi-departmental collaboration. More departmental agents are shown in Appendix \ref{appendix:local_agents}.
\end{enumerate}

Through the above three phases—\emph{plan generation, scoring, and agent allocation}—the planner agent balances exploration and optimization. The beam search mechanism ensures diversity of reasoning, the rubric guarantees safety and logical soundness, and the multi-agent assignment allows personalized, collaborative surgical assistance.

\subsection{Memory Mechanism}
\label{subsection:memory}
A core component of our multi-agent surgical system is the memory mechanism, which draws inspiration from human cognition and recent developments such as MemGPT. The motivation for such a design stems from the observation that patient records exhibit highly skewed length distributions: while some patients have only short clinical notes that can be fully ingested into the model context, others possess long-term hospitalizations with extensive daily progress notes and laboratory test results. Directly inputting the latter would exceed the context window of standard LLMs, necessitating a memory mechanism that supports selective retrieval.

\noindent \textbf{Working Memory.}  
The working memory stores dialogue and reasoning history generated during the current task execution. At each step $t$, when a departmental agent $\mathcal{A}_{dept}$ or the laboratory agent $\mathcal{A}_{lab}$ generates output $o_t$, the system appends this output to the working memory:
\begin{equation}
    \mathcal{M}_{\text{work}}^{(t)} = \mathcal{M}_{\text{work}}^{(t-1)} \cup \{o_t\}.
\end{equation}
This mechanism ensures short-term coherence, allowing subsequent agents to build upon previously generated knowledge.

\noindent \textbf{Long-Term Memory.}  
The long-term memory $\mathcal{M}_{\text{long}}$ maintains structured clinical knowledge from multiple sources:
\begin{equation}
    \mathcal{M}_{\text{long}} = \{ \mathcal{R}_{\text{case}}, \; \mathcal{R}_{\text{record}}, \; \mathcal{R}_{\text{lab}} \},
\end{equation}
where $\mathcal{R}_{\text{case}}$ represents exemplar patient cases containing standardized surgical workflows, $\mathcal{R}_{\text{record}}$ encompasses longitudinal daily medical records, and $\mathcal{R}_{\text{lab}}$ refers to laboratory test results.

The dual-memory design enables the system to maintain short-term coherence across multi-agent reasoning steps, retrieve clinically relevant patient records and exemplar cases as needed, and perform selective abstraction, effectively balancing efficiency with knowledge retention.

\subsection{Department Agents}
\label{subsection:department}
Department agents $\mathcal{A}_{dept}$ are responsible for processing patient information and executing domain-specific tasks. They can be categorized into general clinical departments (e.g., internal medicine, surgery) and laboratory departments. These agents operate in close coordination with both the \emph{working memory} $\mathcal{M}_{\text{work}}$ and the \emph{long-term memory} $\mathcal{M}_{\text{long}}$, thereby ensuring context-aware reasoning and informed decision-making.

\noindent \textbf{General Clinical Departments.}  
For standard clinical departments, the agent retrieves relevant patient historical records and appends concise and important information into the working memory. Specifically, given a patient's basic information $B$, current task description $\tau$, and current surgical step $\pi_t$, the agent generates a retrieval query:
\begin{equation}
    q = f_{\theta}(B, \tau, \pi_t),
\end{equation}
where $f_{\theta}$ is a neural query generator (implemented via an LLM).  
The most relevant record from long-term memory is selected by:
\begin{equation}
    r = \operatorname*{argmax}_{r_i \in \mathcal{M}_{\text{long}}} \text{Sim}(q, r_i),
\end{equation}
where $\text{Sim}(\cdot)$ measures semantic similarity using embeddings. In addition, the agent can identify a set of similar exemplar cases:
\begin{equation}
    \mathcal{C} = \{ c_j \in \mathcal{R}_{\text{case}} \mid \text{Sim}(q, c_j) \geq \delta \},
\end{equation}
where $\delta$ is a predefined similarity threshold. Rather than simply appending retrieved records, each departmental agent $\mathcal{A}_{dept}$ integrates relevant information from long-term memory and generates concise recommendations reflecting its domain expertise and focus. Specifically, the agent output at step $t$ is computed as
\begin{equation}
    o_t^{\mathcal{A}_{dept}} = h_{\psi}^{\mathcal{A}_{dept}} \Big( B, \tau, \pi_t, r^{\mathcal{A}_{dept}}, \mathcal{C}^{\mathcal{A}_{dept}} \Big),
\end{equation}
where $r^{\mathcal{A}_{dept}}$ is the most relevant retrieved record from long-term memory for agent $\mathcal{A}_{dept}$, $\mathcal{C}^{\mathcal{A}_{dept}}$ is the set of cross-referenced exemplar cases, $h_{\psi}^{\mathcal{A}_{dept}}$ denotes the agent-specific reasoning function that produces domain-focused recommendations.

The working memory is then updated with these agent-specific outputs:
\begin{equation}
    \mathcal{M}_{\text{work}}^{(t)} \leftarrow \mathcal{M}_{\text{work}}^{(t-1)} \cup \{ o_t^{\mathcal{A}_{dept}} \mid \mathcal{A}_{dept} \in \mathcal{A}_{\text{local}} \},
\end{equation}
where $\mathcal{A}_{\text{local}}$ is the set of departments involved. 

This ensures that each departmental agent operates with both patient-specific historical context and clinically relevant precedents from comparable cases.

\noindent \textbf{Laboratory Department.}  
For laboratory departments, $\mathcal{A}_{lab}$ performs specialized reasoning over laboratory test results. The workflow is summarized in four stages:

\begin{enumerate}
    \item Abnormality Identification: The system examines all lab tests 
    $L = \{l_1, l_2, \dots, l_m\}$ and identifies abnormal items:
    \begin{equation}
        L_{\text{ab}} = \{ l_i \in L \mid l_i.\text{abnormal} = \text{True} \}.
    \end{equation}
    
    \item Selective Filtering: If the number of abnormal items is large, the agent selects a subset of items most relevant to the patient background $B$ and current task $\tau$:
    \begin{equation}
        L_{\text{sel}} = \text{Select}(L_{\text{ab}}, B, \tau, k_{\max}),
    \end{equation}
    where $k_{\max}$ is the maximum number of items to consider, and $\text{Select}(\cdot)$ ranks items based on clinical relevance and task-specific importance.

    \item Evidence Retrieval and Analysis: For each selected item $l_i \in L_{\text{sel}}$, the agent formulates a query and retrieves relevant literature or guidelines:
    \begin{equation}
        E_i = \text{Retrieve}(l_i.\text{name}, l_i.\text{value}, \text{tools}),
    \end{equation}
    and synthesizes the analysis into structured output $\hat{o}_i$:
    \begin{equation}
        \hat{o}_i = h_{\psi}^{\mathcal{A}_{lab}}(l_i, E_i, B, \tau),
    \end{equation}
    where $h_{\psi}^{\mathcal{A}_{lab}}$ denotes the laboratory agent's reasoning function.
    
    \item Working Memory Update: The synthesized analyses are appended into the working memory:
    \begin{equation}
        \mathcal{M}_{\text{work}}^{(t)} \leftarrow \mathcal{M}_{\text{work}}^{(t-1)} \cup \{ \hat{o}_i \}_{i=1}^{|L_{\text{sel}}|}.
    \end{equation}
\end{enumerate}

\subsection{Aggregation and Task-Level Synthesis}
\label{subsection:aggregation}
Outputs from individual departmental agents are consolidated to form a coherent, task-level summary. This aggregation is performed by the \emph{task-specific agent} $\mathcal{A}_{\text{task}}$, which is designated by the planner to fulfill a particular functional role.

\noindent \textbf{Working Memory Consolidation.}
Let $\mathcal{M}_{\text{work}}^{(t)}$ denote the current working memory after all departmental agents have appended their outputs. The task agent $\mathcal{A}_{\text{task}}$ synthesizes these multi-source contributions into a unified, domain-specific representation:
\begin{equation}
    \hat{o}_{\text{agg}}^{\mathcal{A}_{\text{task}}} = g_{\phi}^{\mathcal{A}_{\text{task}}} \Big( \bigcup_{i=1}^{N} \mathcal{M}_{\text{work},i}^{(t)} \Big),
\end{equation}
where $N$ is the total number of departmental agents involved, $\mathcal{M}_{\text{work},i}^{(t)}$ is the working memory contribution from agent $i$, and $g_{\phi}^{\mathcal{A}_{\text{task}}}$ denotes the summarization function implemented by the task-specific agent. The resulting $\hat{o}_{\text{agg}}^{\mathcal{A}_{\text{task}}}$ provides a concise, structured, and task-focused overview, such as patient analysis, surgical planning, safety considerations, risk assessment, or rehabilitation guidance, depending on the assigned functional role.

\noindent \textbf{Reflection Mechanism.}
To ensure reasoning robustness, the planner performs a \emph{reflective check} on the aggregated output. Let $\mathcal{F}(\cdot)$ denote the reflection operator, which evaluates logical consistency, safety, and completeness:
\begin{equation}
    \hat{o}_{\text{ref}} = \mathcal{F}(\hat{o}_{\text{agg}}),
\end{equation}
where $\hat{o}_{\text{ref}}$ is the revised summary post-reflection. The reflection mechanism allows the system to identify potential conflicts or missing information before finalizing the plan.

\noindent \textbf{Human-in-the-Loop Oversight.}
Despite automated synthesis, clinical expertise remains critical. We incorporate a human-in-the-loop component where a clinician reviews $\hat{o}_{\text{ref}}$ and can provide corrective feedback $\Delta_h$:
\begin{equation}
    \hat{o}_{\text{final}} = \hat{o}_{\text{ref}} + \Delta_h.
\end{equation}
This ensures that the aggregated output adheres to medical standards, captures contextual nuances, and supports accountability in decision-making.

Overall, the aggregation module unifies multi-agent contributions into a coherent summary, applies reflective reasoning for consistency, and integrates human oversight:
\begin{equation}
    \hat{o}_{\text{final}} = \text{HumanReview} \Big( \mathcal{F} \big( g_{\phi} ( \bigcup_i \mathcal{M}_{\text{work},i}^{(t)} ) \big) \Big),
\end{equation}
allowing the system to produce reliable, safe, and clinically actionable guidance while maintaining traceability and accountability.

\begin{algorithm}[H]
\caption{Surgical Plan Generation Algorithm Based on Beam Search}
\label{alg:planner}
\begin{algorithmic}[1]
\Require $max\_steps = 3$, $search\_width = 5$, $beam\_width = 2$
\Ensure Optimal surgical $plan$
    \State $root \gets$ CreateTreeNode($\emptyset$, 0)
    \State $beam \gets [root]$
    \For{$step = 1$ to $max\_steps$}
        \State $candidates \gets \emptyset$
        
        \For{each $parent\_node$ in $beam$}
            \State $next\_plans \gets$ LLM($P$, $B$, $\tau$)
            \For{each $next\_plan$ in $next\_plans$}
                \State $child\_plan \gets parent\_node.plan + [next\_plan]$
                \State $score \gets$ \Call{planner\_evaluator}{$child\_plan$}
                \State $node \gets$ CreateTreeNode($child\_plan$, $score$)
                \State $candidates$.add($node$)
            \EndFor
        \EndFor
        
        \State $beam \gets$ SortByScoreDescending($candidates$)[:$beam\_width$]
    \EndFor
    \State $best\_node \gets beam[0]$
    \State \Return $best\_node.plan$
\end{algorithmic}
\end{algorithm}

\section{Experiments}

\subsection{Experiment Setup}

\noindent \textbf{Dataset.}  
The experimental dataset comprises 530 anonymized, high-quality clinical records collected from top-tier tertiary hospitals. All patient identifiers and sensitive information were removed in compliance with ethical and data protection regulations. The records provide comprehensive perioperative information, including preoperative diagnostics, surgical planning notes, intraoperative safety logs, postoperative complications, and rehabilitation recommendations, enabling end-to-end evaluation of the multi-agent surgical system. The dataset spans patients from five countries, 19 ethnic groups, and 14 occupations, and includes rich metadata (e.g., age, marital status, blood type, and region), ensuring broad applicability across diverse cultural and clinical settings.

\begin{table*}[htbp]
\centering
\caption{Main experimental results across five surgical tasks. Metrics are averaged over three random seeds. Best results are highlighted in bold. ↑ indicates that higher values correspond to better performance, whereas ↓ denotes that lower values are preferable.}
\label{tab:main_results}
\resizebox{\textwidth}{!}{
\begin{tabular}{lllc cccccccc}
\toprule
\multirow{2}{*}{\parbox[c]{2cm}{\centering \textbf{Category}}} &
\multirow{2}{*}{\textbf{Method}} &
\multirow{2}{*}{\textbf{LLM}} &
\multicolumn{2}{c}{\textbf{Analysis}} & 
\multicolumn{2}{c}{\textbf{Surgery}} & 
\multicolumn{2}{c}{\textbf{Safety}} & 
\multicolumn{1}{c}{\textbf{Risk}} & 
\multicolumn{1}{c}{\textbf{Rehab}} \\ 
\cmidrule(lr){4-5} \cmidrule(lr){6-7} \cmidrule(lr){8-9} \cmidrule(lr){10-10} \cmidrule(lr){11-11}
 &  & & DC↑ & MAR↑ & PFS↑ & GAR↑ & EWS↑ & FAR↓ & Recall↑ & Sim↑ \\
\midrule
\multirow{6}{*}{\parbox[c]{2cm}{\centering Vanilla\\Single-agent}} 
 & Zero-shot  & \adjustbox{height=0.35cm, valign=c}{\includegraphics{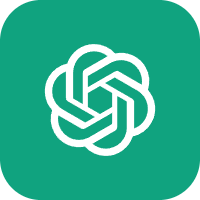}} GPT-4o & 68.4 & 55.7 & 6.32 & 69.2 & 45.1 & 56.3 & 50.5 & 44.6 \\
 & Zero-shot  & \adjustbox{height=0.35cm, valign=c}{\includegraphics{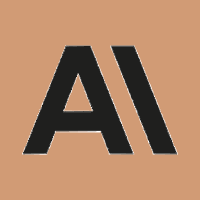}} Claude & 72.1 & 58.8 & 6.57 & 73.4 & 50.3 & 52.8 & 52.4 & 46.1 \\
 & Zero-shot  & \adjustbox{height=0.35cm, valign=c}{\includegraphics{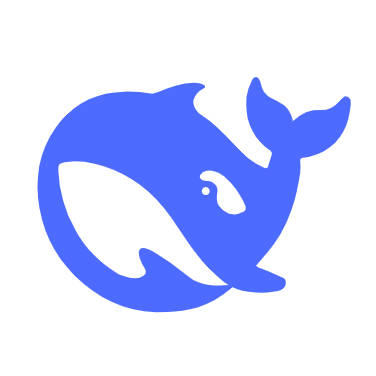}} DeepSeek & 75.6 & 68.4 & 6.84 & 75.1 & 58.2 & 50.7 & 53.2 & 44.5 \\
 & Few-shot   & \adjustbox{height=0.35cm, valign=c}{\includegraphics{chatgpt-logo.png}} GPT-4o & 80.2 & 60.6 & 7.05 & 74.4 & 51.5 & 49.3 & 54.6 & 47.3 \\
 & Few-shot   & \adjustbox{height=0.35cm, valign=c}{\includegraphics{claude-logo.png}} Claude & 83.0 & 64.3 & 7.42 & 78.6 & 55.2 & 46.5 & 55.8 & 49.1 \\
 & Few-shot   & \adjustbox{height=0.35cm, valign=c}{\includegraphics{deepseek-logo.jpg}} DeepSeek & 79.8 & 65.4 & 7.53 & 86.7 & 56.3 & 45.1 & 62.7 & 48.0 \\
\midrule
\multirow{4}{*}{\parbox[c]{2cm}{\centering Advanced\\Single-agent}} 
 & SC \cite{wang2022self}          & \adjustbox{height=0.35cm, valign=c}{\includegraphics{deepseek-logo.jpg}} DeepSeek & 81.3 & 70.2 & 7.94 & 84.5 & 60.2 & 46.3 & 58.2 & 54.3 \\
 & CoT \cite{wei2022chain}         & \adjustbox{height=0.35cm, valign=c}{\includegraphics{deepseek-logo.jpg}} DeepSeek & 82.7 & 71.5 & 7.87 & 83.2 & 61.1 & 45.4 & 59.0 & 55.2 \\
 & CoT-SC      & \adjustbox{height=0.35cm, valign=c}{\includegraphics{deepseek-logo.jpg}} DeepSeek & 78.9 & 73.1 & 8.13 & 86.3 & 63.0 & 43.5 & 60.1 & 56.4 \\
 & MedPrompt \cite{nori2023can}   & \adjustbox{height=0.35cm, valign=c}{\includegraphics{deepseek-logo.jpg}} DeepSeek & 83.5 & 72.4 & 8.27 & 85.4 & 62.3 & 42.8 & 61.2 & 56.1 \\
\midrule
\multirow{5}{*}{\parbox[c]{2cm}{\centering Multi-agent}} 
 & MedAgents \cite{tang2024medagents}            & \adjustbox{height=0.35cm, valign=c}{\includegraphics{deepseek-logo.jpg}} DeepSeek & 88.2 & 75.3 & 8.54 & 91.0 & 65.2 & 40.1 & 63.3 & 58.0 \\
 & ReConcile \cite{chen2024reconcile}            & \adjustbox{height=0.35cm, valign=c}{\includegraphics{deepseek-logo.jpg}} DeepSeek & 87.5 & 78.5 & 8.63 & 89.0 & 64.1 & 38.2 & 65.0 & 60.1 \\
 & MDAgents \cite{kim2024mdagents}             & \adjustbox{height=0.35cm, valign=c}{\includegraphics{deepseek-logo.jpg}} DeepSeek & 86.4 & 77.6 & 8.71 & 90.4 & 67.3 & 37.0 & 66.1 & 61.3 \\
 & ColaCare \cite{wang2025colacare}             & \adjustbox{height=0.35cm, valign=c}{\includegraphics{deepseek-logo.jpg}} DeepSeek & 89.7 & 76.4 & 8.86 & 88.2 & 68.4 & 37.8 & 67.2 & \textbf{62.1} \\
 & \textbf{SURGENT (Ours)} & \adjustbox{height=0.35cm, valign=c}{\includegraphics{deepseek-logo.jpg}} DeepSeek & \textbf{93.1} & \textbf{83.2} & \textbf{9.33} & \textbf{95.3} & \textbf{74.5} & \textbf{33.2} & \textbf{69.4} & 61.0 \\
\bottomrule
\end{tabular}
}
\end{table*}

\noindent \textbf{Baselines.}  
We compare SURGENT against both single-agent and multi-agent baselines.  
\textit{Vanilla Single-agent} include DeepSeek \cite{guo2025deepseek}, GPT-4o \cite{achiam2023gpt} and Claude 3.7-Sonnet \cite{anthropic2024claude3}, each evaluated under both zero-shot and few-shot configurations.  
\textit{Advanced Single-agent} methods further incorporate reasoning-enhanced paradigms, including Self-Consistency (SC) \cite{wang2022self}, Chain-of-Thought (CoT) \cite{wei2022chain}, CoT-SC, and MedPrompt \cite{nori2023can}, all built upon the DeepSeek base model. Specifically, SC samples multiple reasoning paths and selects the most consistent answer to improve reliability; CoT integrates explicit intermediate rationales before deducing the final answer, enhancing interpretability and logical coherence; CoT-SC combines the strengths of CoT and SC by jointly sampling and aggregating multiple rationalized reasoning chains; and MedPrompt composes several prompting strategies to optimize medical reasoning performance, achieving state-of-the-art results across both medical and general-domain benchmarks.  
\textit{Multi-agent} baselines include MedAgents \cite{tang2024medagents}, ReConcile \cite{chen2024reconcile}, MDAgents \cite{kim2024mdagents}, and ColaCare \cite{wang2025colacare}, also using DeepSeek as the foundational backbone.  
This diverse set of baselines ensures comprehensive comparison across reasoning paradigms and collaboration mechanisms.

\noindent \textbf{Evaluation Metrics.}  
Each task is evaluated using domain-specific metrics that reflect its clinical objectives and task characteristics:  
\begin{itemize}
    \item \textbf{Analysis:} Diagnostic Coverage (DC) and Misdiagnosis Avoidance Rate (MAR) are used to assess diagnostic completeness and safety, respectively.  
    \item \textbf{Surgery:} Plan Feasibility Score (PFS) measures the rationality, executability, and safety of generated surgical plans, rated by GPT-4o against reference standards. Guideline Adherence Rate (GAR) quantifies consistency with established surgical guidelines.  
    \item \textbf{Safety:} Early Warning Sensitivity (EWS) evaluates the timeliness of intraoperative risk detection, while False Alarm Rate (FAR) measures the precision of alarm triggering.  
    \item \textbf{Risk:} Complication Recall (Recall) is used to measure the ability to identify all clinically relevant postoperative complications from the reference list.
    \item \textbf{Rehabilitation:} The quality of rehabilitation recommendations is assessed by cosine similarity (Sim) between the generated guidance and reference expert instructions, reflecting both patient-condition alignment and clarity of expression.  
\end{itemize}

\noindent \textbf{Implementation Details.}  
All methods, including the multi-agent frameworks, are implemented with a generation temperature of 0.7. For ToT-based reasoning within the agents, the search depth is set to 3, the search length to 5, and the beam width to 2 at each decoding step. For Vanilla Single-agent and Advanced Single-agent, we add the patient's most relevant medical records and laboratory test results to the prompt. For the tools, we used Tavily Search\footnote{https://www.tavily.com/}, LangChain’s PubMed\footnote{https://python.langchain.com/docs/integrations/retrievers/pubmed/}, and medical guideline retrieval. We used bge-large-zh-v1.5\cite{bge_embedding} as the embedding model due to its strong performance and suitability for Chinese text. Each experiment is repeated three times with different random seeds, and the average performance is reported.

\subsection{Overall Results}

Table~\ref{tab:main_results} presents the main experimental results across five surgical tasks, including analysis, surgery, safety, risk prediction, and rehabilitation. Across the eight evaluation metrics, SURGENT achieves the best performance in seven, consistently outperforming both single-agent and multi-agent baselines.

In the analysis phase, SURGENT achieves the highest DC (93.1\%) and MAR (83.2\%), surpassing the next best method (ColaCare) by 3.4 and 6.8 percentage points, respectively, indicating comprehensive and accurate preoperative assessment capabilities. Among vanilla single-agent, DeepSeek performs best in both zero-shot and few-shot settings, surpassing GPT-4o and Claude across most metrics, though it still falls short of SURGENT.
For the surgical planning metrics, SURGENT leads with PFS of 9.33 and GAR of 95.3\%, reflecting its ability to generate clinically sound and executable surgical plans. This suggests strong practical utility in guiding intraoperative decision-making.
In the safety monitoring phase, SURGENT shows the highest sensitivity (74.5) and lowest FAR (33.2), surpassing ColaCare by 6.1 and 4.6 percentage points, respectively, demonstrating robust awareness of potential complications and superior intraoperative vigilance compared to other baselines.
For postoperative risk detection, SURGENT achieves 69.4\%, exceeding the next best method (MDAgents) by 2.2 percentage points, highlighting its effectiveness in identifying patients at risk of complications and supporting timely intervention.
Finally, in the rehabilitation task, SURGENT provides guidance with a semantic similarity (Sim) of 61.0, slightly below ColaCare (62.1), indicating competitive alignment with expert recommendations and personalized postoperative care planning.

\section{Analysis}

\subsection{Ablation Study of Backbone LLM}

We investigate the influence of different backbone agents on the overall performance of SURGENT. Three widely-used LLMs—GPT-4o, Claude, and DeepSeek—are used as the underlying backbone for SURGENT, while keeping all other configurations identical.

As shown in Table~\ref{tab:ablation_backbone}, the choice of backbone agent has a measurable impact on SURGENT's performance across all surgical intelligence tasks. SURGENT built upon DeepSeek consistently achieves the best performance, surpassing the next best backbone, Claude, by 3.9 percentage points in DC, 0.7 points in MAR, 0.68 points in PFS, and 3.3 percentage points in GAR. The superior performance of DeepSeek can be attributed to its strong reasoning capabilities and advanced Chinese language understanding, which allow it to interpret complex surgical scenarios and generate clinically sound plans.

\begin{table}[htbp]
\centering
\caption{Ablation results for SURGENT with different backbone LLMs. Metrics are averaged over three random seeds. Best results are highlighted in bold.}
\label{tab:ablation_backbone}
\resizebox{\columnwidth}{!}{
\begin{tabular}{lcccc}
\toprule
\textbf{Backbone LLM} & \textbf{DC↑} & \textbf{MAR↑} & \textbf{PFS↑} & \textbf{GAR↑} \\
\midrule
\adjustbox{height=0.35cm, valign=c}{\includegraphics{chatgpt-logo.png}} GPT-4o & 88.4 & 81.7 & 8.51 & 91.2 \\
\adjustbox{height=0.35cm, valign=c}{\includegraphics{claude-logo.png}} Claude & 89.2 & 82.5 & 8.65 & 92.0 \\
\adjustbox{height=0.35cm, valign=c}{\includegraphics{deepseek-logo.jpg}} DeepSeek & \textbf{93.1} & \textbf{83.2} & \textbf{9.33} & \textbf{95.3} \\
\midrule
 & \textbf{EWS↑} & \textbf{FAR↓} & \textbf{Recall↑} & \textbf{Sim↑} \\
\adjustbox{height=0.35cm, valign=c}{\includegraphics{chatgpt-logo.png}} GPT-4o & 70.3 & 36.8 & 66.1 & 57.4 \\
\adjustbox{height=0.35cm, valign=c}{\includegraphics{claude-logo.png}} Claude & 72.1 & 35.5 & 67.5 & 58.2 \\
\adjustbox{height=0.35cm, valign=c}{\includegraphics{deepseek-logo.jpg}} DeepSeek & \textbf{74.5} & \textbf{33.2} & \textbf{69.4} & \textbf{61.0} \\
\bottomrule
\end{tabular}
}
\end{table}

This ablation study demonstrates the critical role of selecting a capable backbone LLM in multi-agent surgical intelligence systems. In particular, DeepSeek not only delivers superior performance across analysis, planning, safety, risk prediction, and rehabilitation tasks, but is also open-source and can be deployed privately, enabling broader and more flexible clinical applications. Moreover, all backbone models perform well within our multi-agent framework, highlighting the strong generalization capability of the system.

\subsection{Ablation Study of Components}
\begin{table}[htbp]
\centering
\caption{Ablation study of SURGENT system components. Metrics are averaged over three random seeds. Best results are highlighted in bold.}
\label{tab:ablation_components}
\resizebox{\columnwidth}{!}{
\begin{tabular}{lcccc}
\toprule
\textbf{System Variant} & \textbf{DC↑} & \textbf{MAR↑} & \textbf{PFS↑} & \textbf{GAR↑} \\
\midrule
Full SURGENT & \textbf{93.1} & \textbf{83.2} & \textbf{9.33} & \textbf{95.3} \\
w/o Planner Agent             & 90.2 & 81.5 & 8.90 & 92.8 \\
w/o Memory Mechanism          & 85.7 & 77.8 & 8.24 & 88.7 \\
w/o Department Agents         & 82.5 & 74.3 & 7.87 & 85.4 \\
w/o Aggregation Module        & 89.5 & 80.9 & 8.72 & 92.1 \\
\midrule
 & \textbf{EWS↑} & \textbf{FAR↓} & \textbf{Recall↑} & \textbf{Sim↑} \\
Full SURGENT & \textbf{74.5} & \textbf{33.2} & \textbf{69.4} & \textbf{61.0} \\
w/o Planner Agent             & 71.3 & 36.2 & 66.7 & 58.1 \\
w/o Memory Mechanism          & 65.8 & 40.7 & 62.3 & 55.4 \\
w/o Department Agents         & 60.5 & 44.8 & 58.2 & 50.6 \\
w/o Aggregation Module        & 70.4 & 35.6 & 66.3 & 57.2 \\
\bottomrule
\end{tabular}
}
\end{table}

To better understand the contribution of each component in the SURGENT system, we conducted an ablation study by systematically removing individual modules and evaluating the system performance across multiple surgical tasks. The results are summarized in Table~\ref{tab:ablation_components}.  

Ablation studies reveal the contributions of different components: without the planner agent, surgical plans are generated directly from task descriptions, resulting in a moderate performance decline (DC drops from 93.1\% to 90.2\%) due to less structured planning; removing the memory mechanism, which selects the most relevant patient records and test results, causes a more substantial decrease (DC to 85.7\%, MAR to 77.8\%), highlighting the critical role of long-term memory in tailoring decisions to patient-specific information; omitting department agents, so that all doctors operate without specialty-based division, further reduces task-specific reasoning capabilities, leading to significant performance loss (DC to 82.5\%, MAR to 74.3\%) and illustrating the importance of department-specific expertise for precise multi-agent coordination. Finally, replacing the task-specific aggregation module with a unified aggregation strategy moderately degrades performance (DC to 89.5\%, MAR to 80.9\%), indicating that task-adaptive aggregation effectively integrates outputs from multiple agents. From the table, it is evident that the removal of the Memory Mechanism and Department Agents has the most significant impact on the system.

\subsection{Ablation Study of ToT}
\begin{figure}
    \centering
    \includegraphics[width=\linewidth]{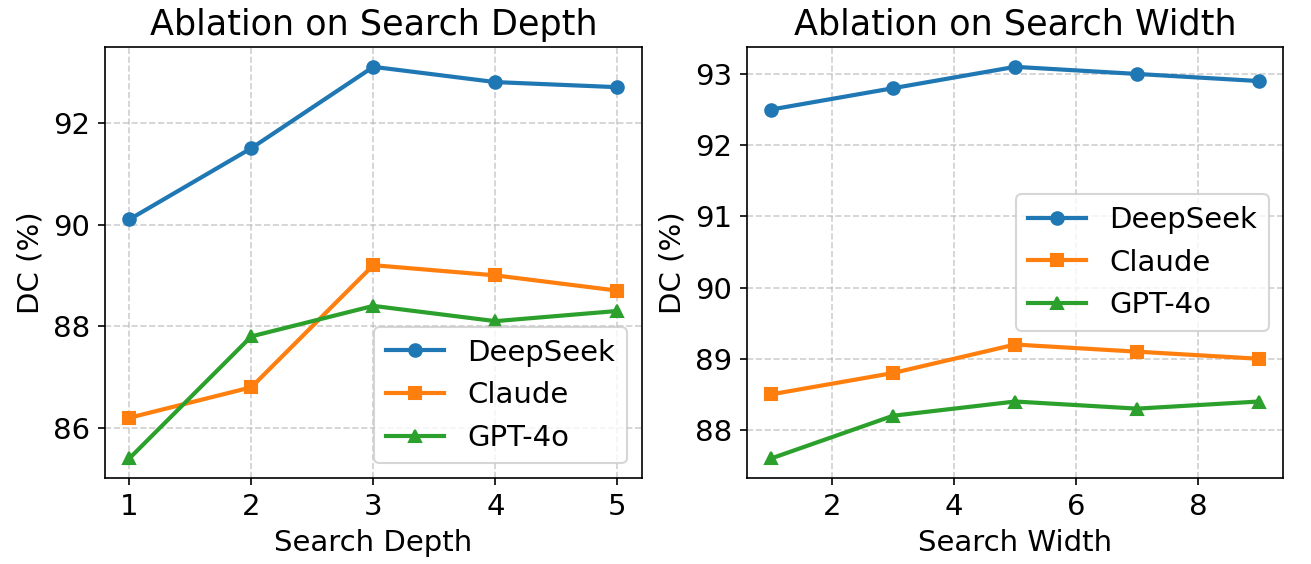}
    \caption{Ablation Study of ToT}
    \label{fig:Ablation Study of ToT}
\end{figure}

To investigate the influence of the ToT search parameters in the planner agent, we conduct ablation experiments on both \textit{search depth} and \textit{beam width}. As illustrated in Figure~\ref{fig:Ablation Study of ToT}, the performance (measured by DC) consistently improves as the search depth or width increases, reaching the best results at depth = 3 and width = 5. 

When the search depth is set to 1, the planner agent effectively degenerates into a single-step generator without structured reasoning or multi-step planning, leading to a noticeable drop in accuracy (e.g., DC decreases from 93.1\% to 90.1\%).  However, further increasing depth beyond 3 introduces redundant exploration and unstable intermediate decisions, causing slight performance degradation (e.g., 93.1\% $\rightarrow$ 92.7\%).

A similar trend is observed for beam width. The system benefits from broader candidate exploration up to width = 5, beyond which the performance saturates or slightly declines (e.g., 93.1\% $\rightarrow$ 92.9\%), likely due to the inclusion of noisy or low-quality plans. 

\subsection{Rationality of Collaboration}

\begin{table}[htbp]
\centering
\caption{Expert evaluation of the rationality of department agents and inter-agent collaboration. Scores are averaged over 5 expert responses (1–5 scale).}
\label{tab:expert_rationality}
\resizebox{0.95\linewidth}{!}{
\begin{tabular}{lcc}
\toprule
\textbf{Department Agent} & \textbf{Individual↑} & \textbf{Collaboration↑} \\
\midrule
Department Agents & 4.65 & 4.70 \\
Laboratory Agent & 4.58 & 4.63 \\
\bottomrule
\end{tabular}
}
\end{table}

To evaluate the rationality of the surgical multi-agent collaboration framework, we invited five domain experts to assess the system’s performance in generating surgical plans. The scoring criteria are shown in Appendix \ref{appendix:Scoring Criteria for Expert Evaluation}. Each expert rated the system on a 1–5 scale in two dimensions:  
(1) \textit{Individual Rationality}—the extent to which each departmental agent’s reasoning and recommendations reflect realistic clinical judgment, and  
(2) \textit{Collaboration Rationality}—the degree to which inter-departmental coordination resembles actual multidisciplinary teamwork in surgery.

As presented in Table~\ref{tab:expert_rationality}, all department agents achieved high ratings in both dimensions. These results indicate that each agent can produce clinically sound decisions individually, while the overall collaboration mechanism effectively reproduces the rational and coordinated workflow expected in real surgical planning scenarios.

\subsection{Comprehensiveness of Memory}
To assess whether the proposed dual-memory architecture effectively mitigates incomplete or inconsistent reasoning, we conducted an expert evaluation focusing on the comprehensiveness and accuracy of the generated clinical reasoning outputs. Five domain experts reviewed the case analyses generated by the multi-agent system for 20 patient cases, each containing complete daily records. Experts rated the results on a 1–5 scale along two dimensions: (1) \textit{Comprehensiveness}—whether the system captured all key symptoms, findings, and temporal progressions; and (2) \textit{Accuracy}—whether the reasoning process reflected clinically correct interpretations and priorities.

The dual-memory mechanism received an average score of 4.68 for comprehensiveness and 4.69 for accuracy, indicating a high level of agreement among experts regarding its effectiveness. Experts observed that combining short-term task context ($\mathcal{M}{\text{work}}$) with long-term patient information ($\mathcal{M}{\text{long}}$) allowed the system to maintain reasoning consistency across multi-step tasks.

\section{System Overhead}
During the planner agent stage, due to the need for continuous evaluation and iteration of the plan, it takes approximately 20 seconds, consuming about 2100 input tokens and 150 output tokens. The total system time is around 30 seconds, with a total token consumption of 5300, balancing efficiency and system reliability. For surgical planning and post-operative guidance, longer delays are acceptable. However, for intraoperative detection, we run the high-cost inference processes in advance, and then perform inference based on the patient's real-time data to generate a response within 3 seconds.

\section{Related Work}
\subsection{Multi-Agent Collaboration} 
An LLM agent is a system powered by a LLM that can autonomously pursue goals, performing tasks such as reasoning, planning, and managing memory \cite{chang2024agentboard}. Multi-agent collaboration utilizes several agents working together to address complex problems that a single agent might not be able to solve alone. General-purpose collaborative agent frameworks, such as AutoGPT \cite{autogpt2023}, XAgent \cite{xagent2023}, and MetaGPT \cite{hong2024metagpt}, have shown the effectiveness of breaking down tasks and assigning specialized roles to individual agents. At the heart of multi-agent collaboration frameworks is the design of their collaboration mechanisms. For example, MDAgents \cite{kim2024mdagents} classifies medical queries by complexity, recruits corresponding experts, performs reasoning or multi-agent discussion; MedAgents \cite{tang2024medagents}allows LLM-based experts to collaborate in five steps—gather, analyze, summarize, consult; ReConcile \cite{chen2024reconcile} simulates roundtable discussions where diverse LLMs generate initial answers with reasoning and confidence, iteratively debate using past responses and human-like explanations; voting mechanisms integrate the outputs of individual agents—often weighted by their confidence scores—to reach a consensus decision \cite{wang2022self}, while debate strategies enable agents to iteratively challenge and refine each other’s reasoning, improving the quality and robustness of the final solution \cite{du2023improving}; ColaCare \cite{wang2025colacare} simulates multi-disciplinary team discussions to coordinate complex decision-making. 

\subsection{AI in Surgical Domains}
Early research on surgical intelligence primarily focused on tool detection and phase recognition from intraoperative videos, aiming to capture procedural workflows and instrument usage patterns \cite{twinanda2016single, czempiel2020tecno}. With the rise of transformer-based architectures, subsequent works such as TeCNO \cite{czempiel2020tecno} and Trans-SVNet \cite{gao2021trans} improved temporal reasoning by modeling long-range dependencies across surgical phases. However, these methods lack the capability for reasoning, planning, and adaptive decision-making in complex surgical contexts.
Recent studies have begun exploring LLMs for clinical and surgical reasoning. Systems such as Med-PaLM \cite{singhal2025toward} and BioGPT \cite{luo2022biogpt}demonstrated the potential of LLMs in structured medical question answering and clinical text summarization. SurgeryLLM \cite{ong2024surgeryllm} uses RAG to retrieve external knowledge (e.g., clinical guidelines) to enhance reliability. 

\section{Conclusion}
In this paper, we present SURGENT, a surgical multi-agent assistance system that supports intelligent decision-making across the perioperative workflow. Integrating a Tree-of-Thought planner, multi-department agents, and a dual-memory mechanism for both long- and short-term reasoning, SURGENT overcomes key limitations of web-based LLMs in surgical contexts by enabling transparent, traceable, and privacy-preserving reasoning.
Experiments on five perioperative tasks—analysis, planning, safety monitoring, risk assessment, and rehabilitation—show that SURGENT outperforms baseline LLMs and multi-agent systems, with ablation studies confirming the robustness of DeepSeek as its backbone. SURGENT thus advances reliable and equitable AI assistance for surgical care, with future work focusing on broader clinical adaptation and real-time system integration.

\section*{Ethics Statement}
Our project was conducted in collaboration with top-tier tertiary hospitals in China, with proper approval for all data sharing. All patient data used in this study were fully de-identified to ensure privacy and confidentiality, and no individual patient or medical personnel can be identified from the dataset. The study protocol was reviewed and approved by the institutional ethics committees of the participating hospitals. Due to legal restrictions, the data are available for research purposes only; researchers may contact us with their research objectives and intended use. We ensure full compliance with applicable laws and ethical guidelines during data collection and use.

\newpage

\bibliographystyle{ACM-Reference-Format}
\bibliography{main}

\appendix

\section{Scoring Criteria for Expert Evaluation}

\label{appendix:Scoring Criteria for Expert Evaluation}
To ensure a consistent and interpretable assessment of the proposed surgical multi-agent collaboration framework, experts evaluated each dimension—\textit{Individual Rationality} and \textit{Collaboration Rationality}—on a 1–5 scale according to the following criteria:

\begin{itemize}
    \item \textbf{Individual Rationality:} Evaluates whether a single departmental agent’s reasoning and recommendations are clinically sound, consistent with professional guidelines, and logically coherent.
    \begin{itemize}
        \item \textbf{5 – Excellent:} Fully aligns with realistic clinical reasoning; decisions are precise, evidence-based, and context-aware.  
        \item \textbf{4 – Good:} Generally consistent with clinical logic, with only minor omissions or simplifications.  
        \item \textbf{3 – Fair:} Reasoning is partially appropriate but includes noticeable gaps or inconsistencies.  
        \item \textbf{2 – Poor:} Reasoning shows limited clinical relevance or flawed logical progression.  
        \item \textbf{1 – Unacceptable:} Decisions are clinically implausible or irrelevant.  
    \end{itemize}

    \item \textbf{Collaboration Rationality:} Assesses how well different departmental agents coordinate and communicate to produce integrated surgical decisions.
    \begin{itemize}
        \item \textbf{5 – Excellent:} Collaboration closely mirrors real multidisciplinary teamwork; information flow and task division are efficient and complementary.  
        \item \textbf{4 – Good:} Coordination is clear and coherent, though with minor inefficiencies or overlaps.  
        \item \textbf{3 – Fair:} Basic collaboration occurs, but with limited integration or occasional misunderstandings between agents.  
        \item \textbf{2 – Poor:} Agents work mostly independently with little meaningful interaction.  
        \item \textbf{1 – Unacceptable:} Collaboration is absent or counterproductive, leading to incoherent outcomes.  
    \end{itemize}
\end{itemize}

These criteria ensure that both the individual reasoning capability and the collective coordination quality of the multi-agent system are quantitatively and qualitatively assessed.

\section{Plan Feasibility Score (PFS) Criteria}

\textbf{Objective:} To comprehensively evaluate surgical plans in terms of personalization, rationality, and safety, ensuring that the proposed plan is clinically executable and tailored to the patient.

\subsection{Scoring Dimensions}

\begin{enumerate}
    \item \textbf{Personalization (P)} \\
    Definition: Whether the plan sufficiently considers the patient's specific condition, medical history, and individual characteristics. \\
    Example Scores:
    \begin{itemize}
        \item 1.0: Highly tailored to the patient, fully meets individual needs
        \item 0.5: Partially considers patient-specific factors, room for improvement
        \item 0.0: Does not reflect individual patient characteristics
    \end{itemize}

    \item \textbf{Rationality (R)} \\
    Definition: Logical sequence of surgical steps, procedural workflow, and clinical feasibility. \\
    Example Scores:
    \begin{itemize}
        \item 1.0: Steps are complete, logically clear, and comply with clinical norms
        \item 0.5: Minor issues in sequence or logic, but overall feasible
        \item 0.0: Steps are illogical or violate clinical standards
    \end{itemize}

    \item \textbf{Safety (S)} \\
    Definition: The plan's ability to minimize patient risks, including complication prevention and critical operation safety. \\
    Example Scores:
    \begin{itemize}
        \item 1.0: Minimizes risk to the greatest extent, meets safety standards
        \item 0.5: Some risks exist but are manageable
        \item 0.0: Unacceptable risks present
    \end{itemize}
\end{enumerate}

\subsection{Calculation Method}
\[
\text{PFS} = w_P \cdot P + w_R \cdot R + w_S \cdot S
\]

Where $w_P$, $w_R$, and $w_S$ are weights reflecting the relative importance of each dimension (default: 1/3 each).

\section{Evaluation Metrics}

This appendix formally defines the evaluation metrics used to assess the performance of the SURGENT multi-agent system across different clinical tasks.

\subsection{Case Analysis}

\begin{itemize}
    \item \textbf{Diagnostic Coverage (DC)}
    \begin{equation}
        DC = \frac{|\text{Detected Diagnoses}|}{|\text{Reference Diagnoses}|}
    \end{equation}
    Measures the proportion of clinically relevant diagnoses correctly identified by the agent.
    
    \item \textbf{Misdiagnosis Avoidance Rate (MAR)}
    \begin{equation}
        MAR = \frac{|\text{Avoided Misdiagnoses}|}{|\text{Potential Misdiagnoses}|}
    \end{equation}
    Reflects the agent’s effectiveness in reducing misdiagnoses or missed diagnoses.
\end{itemize}

\subsection{Surgical Plan Generation}

\begin{itemize}
    \item \textbf{Plan Feasibility Score (PFS)}
    \begin{equation}
        PFS \in [0,1]
    \end{equation}
    Rated by expert reference (or GPT-4o) based on step rationality, executability, and safety.
    
    \item \textbf{Guideline Adherence Rate (GAR)}
    \begin{equation}
        GAR = \frac{|\text{Steps aligned with guidelines}|}{|\text{All Steps}|}
    \end{equation}
    Measures the proportion of steps in the generated plan that comply with established surgical guidelines.
\end{itemize}

\subsection{Intraoperative Safety Monitoring}

\begin{itemize}
    \item \textbf{Early Warning Sensitivity (EWS)}
    \begin{equation}
        EWS = \frac{TP}{TP + FN}
    \end{equation}
    Quantifies the proportion of true risk events detected in time during surgery.
    
    \item \textbf{False Alarm Rate (FAR)}
    \begin{equation}
        FAR = \frac{FP}{TP + FP}
    \end{equation}
    Measures the proportion of incorrectly triggered alerts to avoid unnecessary interruptions.
\end{itemize}

\subsection{Postoperative Complication Risk}

\begin{equation}
    Recall = \frac{|\text{Detected Complications}|}{|\text{Reference Complications}|}
\end{equation}
Reflects the ability to identify all clinically relevant postoperative complications from the reference list.

\subsection{Rehabilitation Recommendation}

\begin{equation}
    Sim = \frac{\mathbf{v}_{\text{generated}} \cdot \mathbf{v}_{\text{reference}}}
                {\|\mathbf{v}_{\text{generated}}\| \|\mathbf{v}_{\text{reference}}\|}
\end{equation}
Cosine similarity between generated guidance and reference expert instructions, measuring both alignment with patient condition and clarity of expression.

\section{Expert Rating Scale for Memory Comprehensiveness and Accuracy}

To systematically evaluate the comprehensiveness and accuracy of clinical reasoning outputs generated by the multi-agent system, a 1–5 rating scale was used by domain experts. The scale is defined as follows:

\begin{itemize}
    \item \textbf{1 – Very Poor:} The output captures very few key symptoms, findings, or temporal progressions, and contains multiple clinically incorrect interpretations or priorities.
    \item \textbf{2 – Poor:} The output captures some relevant information but misses important elements, with noticeable errors in reasoning or prioritization.
    \item \textbf{3 – Fair:} The output captures most key elements but may have minor omissions or occasional inaccuracies in clinical interpretation.
    \item \textbf{4 – Good:} The output captures almost all relevant symptoms, findings, and progressions, with reasoning largely consistent with clinical expectations; minor inaccuracies may exist.
    \item \textbf{5 – Excellent:} The output is complete and fully accurate, capturing all key clinical information and maintaining correct reasoning and prioritization throughout.
\end{itemize}

\section{Department Agents}
\label{appendix:local_agents}

\begin{table}[h]
\centering
\caption{Department agents in our proposed surgical multi-agent assistance system}
\label{tab:department_agents}
\begin{tabular}{ll}
\toprule
Department & Agent \\
\midrule
Respiratory Medicine       & respiratory\_medicine\_agent \\
Cardiovascular Medicine    & cardiovascular\_medicine\_agent \\
Neurology                  & neurology\_agent \\
General Surgery            & general\_surgery\_agent \\
Cardiothoracic Surgery     & cardiothoracic\_surgery\_agent \\
Neurosurgery               & neurosurgery\_agent \\
Orthopedics                & orthopedics\_agent \\
Urology                    & urology\_agent \\
Obstetrics \& Gynecology   & ob\_gyn\_agent \\
Pediatrics                 & pediatrics\_agent \\
ENT                        & ent\_agent \\
Emergency Medicine         & emergency\_agent \\
ICU                        & icu\_agent \\
Pathology                  & pathology\_agent \\
Anesthesiology             & anesthesiology\_agent \\
Oncology                   & oncology\_agent \\
Rehabilitation             & rehabilitation\_agent \\
Preventive Healthcare      & preventive\_healthcare\_agent \\
\bottomrule
\end{tabular}
\end{table}

\end{document}